\documentclass[3p,times,procedia,preprint]{elsarticle}
\usepackage[bookmarks=false]{hyperref}
    \hypersetup{colorlinks,
      linkcolor=blue,
      citecolor=blue,
      urlcolor=blue}

\usepackage{amsmath,amsfonts,amssymb}
\usepackage{algorithmic}
\usepackage{algorithm}
\usepackage{array}
\usepackage[caption=false,font=normalsize,labelfont=sf,textfont=sf]{subfig}
\usepackage{url}
\usepackage{graphicx}
\usepackage[utf8]{inputenc}
\usepackage{amsthm}
\usepackage{booktabs}
\usepackage{bbding}
\usepackage[dvipsnames]{xcolor}
\usepackage{multirow}
\usepackage{pifont}

\newcommand{\cmark}{\textcolor{ForestGreen}{\checkmark}}
\newcommand{\xmark}{\textcolor{Red}{\ding{55}}}

\usepackage{amssymb}
\usepackage[figuresright]{rotating}
\begin{document}
\begin{frontmatter}



% \dochead{30th International Conference on Knowledge-Based and Intelligent Information \& Engineering Systems (KES 2026)}%
%% e.g. \dochead{17th International Conference on Dynamical Processes in Excited States of Solids}

\title{Audio Sentiment Analysis via Distillation and Cross-Modal Integration of Generated Multilingual Transcripts}


\author[a]{Andrei-George Durdun} 
\author[b]{Victor Constantinescu}
\author[a,b]{Radu Tudor Ionescu\corref{cor1}}

\address[a]{Department of Computer Science, University of Bucharest, 15G Iuliu Maniu, Bucharest 061075, Romania}
\address[b]{Department of Data Science, PPC Romania, 30 Mircea Voda, Bucharest 030667, Romania}

\begin{abstract}
Automatically recognizing the sentiment, positive or negative, from speech is a challenging task, requiring both the analysis of vocal inflections and the interpretation of uttered words. Recent solutions rely on audio foundation models to solve the task, but it remains unclear if such models can take all aspects into account. To this end, we propose a multimodal solution that integrates audio and text information via cross-modal transformers, where text transcripts are automatically generated via an automatic speech recognition (ASR) tool. Moreover, we create multiple text modalities by automatically translating the transcripts into multiple languages via machine translation tools. Audio and multilingual text features are combined via a cascaded architecture comprising cross-modal transformer blocks that integrate modalities one by one. We further distill knowledge from the multimodal model, called teacher, into a unimodal (audio only) model, called student. We conduct experiments on a large-scale dataset, demonstrating that the automatically generated textual information can bring significant performance boosts in multimodal sentiment polarity classification. Our ablation study confirms that both automatic transcripts and automatic translations are helpful. Moreover, we show that the audio-only model can be enhanced via distillation, boosting performance without any computational overhead during inference. To reproduce the reported results, we publicly release our code at \url{https://github.com/andreidurdun/cross-modal-audio-sentiment}.
\end{abstract}

\cortext[cor1]{Corresponding author.}
\end{frontmatter}

%\correspondingauthor[*]{Corresponding author. Tel.: +0-000-000-0000 ; fax: +0-000-000-0000.}
\email{radu.ionescu@fmi.unibuc.ro}

% \linenumbers

\setlength{\abovedisplayskip}{3.8pt}
\setlength{\belowdisplayskip}{3.8pt}
\setlength{\abovedisplayshortskip}{3pt}
\setlength{\belowdisplayshortskip}{3pt}

\vspace{-0.1cm}
\section{Introduction}
\vspace{-0.1cm}
Sentiment polarity classification from speech is an actively studied task \cite{Atmaja-Sen-2022,Garcia-ICTC-2024,Luo-AffCon-2019,Mohanty-ADMTMSC-2022,Shruti-BigComp-2023}, having a broad range of real-world applications, such as customer service call analysis \cite{Bulkrock-ICTCS-2025}, virtual assistant adaptation \cite{Anilsagar-JSEE-2025}, mental health monitoring \cite{Shanthi-HTL-2025}, vehicle driver monitoring \cite{Zhang-TAC-2026} and gaming experience adaptation \cite{Liu-EC-2025}, among others. As for most speech processing tasks nowadays, deep learning models \cite{Gong-INTERSPEECH-2021,Kim-ICML-2021,Pratap-JMLR-2024,Ristea-INTERSPEECH-2022}, especially audio foundation models \cite{Baevski-NeurIPS-2020,Chen-JSTSP-2022,Pratap-JMLR-2024,Radford-ICML-2023}, represent the mainstream solution for polarity classification, due to their typically high accuracy levels. However, the task remains challenging, as it involves the concurrent analysis of several aspects, including tone (which transmits emotion), pitch (which conveys excitement or calmness), word usage (which reflects the polarity of the transmitted message), irony / sarcasm (which may often indicate the opposite opinion than the transmitted message). Disentangling these aspects is a key pathway towards improving performance, yet this area remains largely underexplored in current literature.

\begin{figure}[t]
  \centering
  \includegraphics[width=0.82\linewidth]{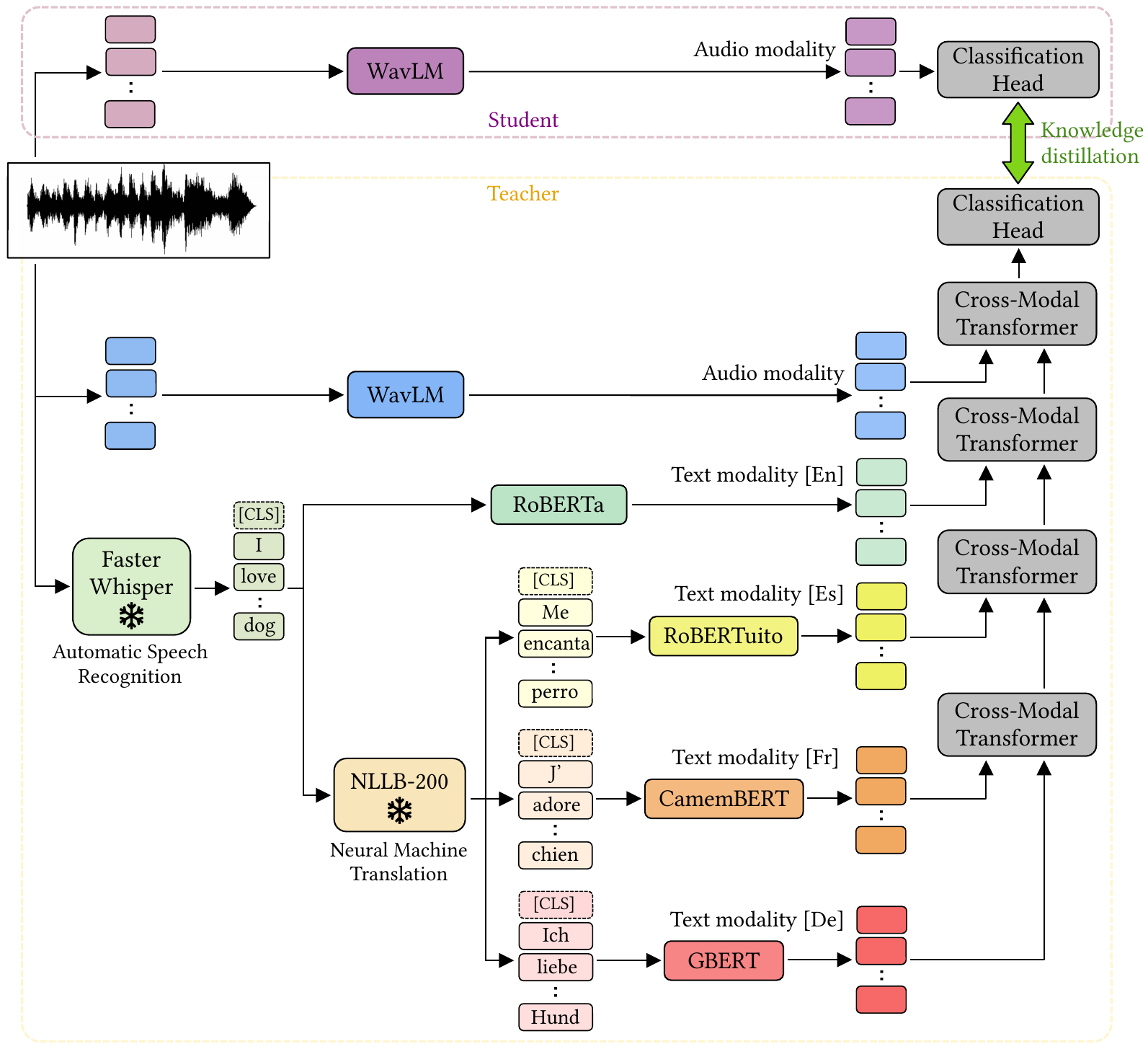}
  \vspace{-0.2cm}
  \caption{The proposed pipeline based on learning under privileged information, which distills information from a multimodal (audio-text) teacher model into a unimodal (audio only) student model. The teacher model comprises frozen ASR and NMT models to generate text transcripts and translations, respectively. Trainable models process the generated texts to extract latent representations that are further integrated via cross-modal transformer blocks. Best viewed in color.}
  \label{fig:pipeline}
  \vspace{-0.3cm}
\end{figure}

To this end, we propose a novel knowledge distillation (KD) pipeline~\cite{Hinton-NIPS-2015}, where the teacher model benefits from disentangled audio and text information. While the audio modality is readily available, the text transcript is not directly accessible in real-world scenarios. We therefore employ an automatic speech recognition (ASR) model to generate text transcripts. However, ASR technology is not perfect, and consequently, the transcripts may be affected by recognition errors. Moreover, language-specific pre-trained models, e.g.~BERT \cite{Devlin-NAACL-2019}, may be affected by training dataset biases. To mitigate biases of language-specific models, we further propose to employ a neural machine translation (NMT) system to automatically translate ASR transcripts into multiple languages, namely German (DE), French (FR) and Spanish (ES). The resulting text modalities along with the original audio modality are integrated into a unified model based on cascaded cross-modal transformer blocks, as illustrated in Figure \ref{fig:pipeline}. Each cross-modal transformer integrates one modality with the upstream information, until all modalities are combined into the joint model. 

While the ASR and NMT models provide the means to disentangle voice and message information, they also represent a processing bottleneck, i.e.~generating the text transcripts and translations requires extra computational resources and processing time. The processing speed is of utter importance for applications requiring real-time sentiment polarity assessment, such as virtual assistant adaptation and vehicle driver monitoring. To harness the multimodal information while preserving the processing time of the unimodal (audio only) model, we employ learning under privileged information (LUPI) \cite{LopezPaz-ICLR-2016,Menadil-ICPR-2024,Vapnik-JMLR-2015}, a knowledge distillation (KD) pipeline that transfers knowledge learned by a teacher model having access to extra data representations (in our case, the generated text modalities) to a student model which can only access the original representation (in our case, the audio modality). The main advantage of this distillation framework is that the extra modalities are not needed during inference. This means that the ASR, the NMT and the multiple text embedding models are only used at training time. The resulting student model represents an effective and scalable solution for audio sentiment classification.

In summary, our framework relies on two assumptions: (i) generated transcripts and translations provide additional sentiment polarity clues that can be effectively harnessed by a multimodal model; (ii) knowledge from the heavy multimodal teacher model can be distilled into an efficient unimodal student. To validate the two hypotheses, we conduct experiments on the large-scale MSP-Podcast corpus \cite{busso2025msp}. Our experiments show that integrating generated multimodal information leads to significant performance gains, up to $+5.89\%$ in terms of macro-$F_1$ and $+5.15\%$ in terms of accuracy, confirming hypothesis (i). Moreover, distilling knowledge from the multimodal teacher into the audio-only student boosts the performance of the student by $+1.54\%$ in terms of macro-$F_1$ and $+0.81\%$ in terms of accuracy, thus validating hypothesis (ii). Furthermore, we show ablation results justifying the utility of both text transcripts and translations.

In summary, our contribution is twofold:
\begin{itemize}
    \item \vspace{-0.2cm} We propose a novel knowledge distillation pipeline for audio sentiment classification, which harnesses both generated text transcripts and translations via a cascaded cross-modal transformer architecture able to effectively integrate multiple modalities.  
    \item We perform experiments showing that generated text modalities lead to considerable performance gains, and that knowledge from a multimodal teacher model can be effectively transferred to an efficient unimodal student.
\end{itemize}

\vspace{-0.3cm}
\section{Related Work}
\label{sec:related}
\vspace{-0.1cm}
 
\noindent
\textbf{Sentiment/emotion recognition from speech.}
Sentiment polarity classification from speech \cite{Atmaja-Sen-2022,Garcia-ICTC-2024,Luo-AffCon-2019,Mohanty-ADMTMSC-2022,Shruti-BigComp-2023}, a.k.a.~audio sentiment analysis, is the task of automatically determining the polarity of the sentiment verbally expressed by a person, where typical options are positive, neutral and negative. The task is deeply connected to speech emotion recognition (SER) \cite{GEORGE2024127015,Lian-Entropy-2023}, where the aim is to precisely establish the emotion class, e.g.~happy, anger or fear, going beyond polarity classification. Both tasks have evolved from hand-crafted acoustic features and conventional classifiers towards end-to-end deep models operating directly on raw waveforms \cite{GEORGE2024127015,Lian-Entropy-2023}. Early deep learning solutions relied on convolutional and recurrent architectures applied to spectrograms or Mel-frequency cepstral coefficients, with attention-based pooling becoming a common strategy for aggregating frame-level emotional cues into utterance-level predictions \cite{GEORGE2024127015}. More recently, the dominant paradigm has shifted towards large self-supervised speech foundation models, such as wav2vec~2.0 \cite{Baevski-NeurIPS-2020}, HuBERT \cite{Hsu-TASLP-2021} and WavLM \cite{Chen-JSTSP-2022}, which provide universal representations that transfer well to paralinguistic tasks. Wang et al.~\cite{Wang-ARXIV-2021} benchmarked wav2vec~2.0 and HuBERT for SER, speaker verification and spoken language understanding, showing that fine-tuning self-supervised encoders consistently outperforms spectrogram-based baselines. WavLM, in particular, jointly models masked speech prediction and denoising, and has emerged as a strong backbone for SER in naturalistic conditions \cite{busso2025msp,Chen-JSTSP-2022}.
 
Progress in naturalistic SER has been largely driven by the MSP-Podcast corpus \cite{busso2025msp}, a large-scale collection of spontaneous emotional speech that supports both categorical and attribute-based tasks. The INTERSPEECH 2025 Speech Emotion Recognition Challenge \cite{Naini-Interspeech-2025} further confirmed the difficulty of the task in the wild: across 93 participating teams, top-performing systems combined audio foundation models with text-based features obtained via automatic speech recognition (ASR), highlighting the practical value of multimodal supervision when reference transcripts are unavailable. While these approaches achieve strong results, most submissions adopt heavy multimodal inference pipelines, raising a natural question about whether part of this gain can be compressed back into audio-only deployments. Our work adopts MSP-Podcast and WavLM as the basis for an audio-centered pipeline, and directly targets this efficiency question via distillation.
 
\noindent
\textbf{Multimodal audio-text sentiment/emotion analysis.}
A complementary line of research jointly models acoustic and linguistic information, motivated by the observation that prosody and lexical content carry largely complementary affective cues \cite{Lian-Entropy-2023,Yoon-SLT-2018}. Yoon et al.~\cite{Yoon-SLT-2018} proposed one of the early audio-text dual-encoder architectures for emotion recognition, combining an RNN over acoustic features with a textual encoder. Tsai et al.~\cite{Tsai-ACL-2019} introduced the Multimodal Transformer (MulT), in which directional cross-modal attention aligns unaligned audio, text and visual streams, establishing cross-modal attention as the mainstream fusion mechanism. %for CMU-MOSI and CMU-MOSEI. 
Subsequent work has explored alternative fusion strategies, including multi-scale locally aggregated descriptors \cite{Luo-ARXIV-2022}, self-adjusting cross-modal representations that preserve unimodal characteristics \cite{Zhang-ARXIV-2022}, and hierarchical cross-modal attention with dual audio pathways \cite{Vamsidhar-SciRep-2025}. Ristea et al.~\cite{Ristea-AIRE-2024,Ristea-ACMMM-2023} further proposed a cascaded cross-modal transformer (CCMT) that fuses multiple textual views derived from different translations of the same utterance with the audio signal, which directly inspired the fusion module used by our teacher.
 
A practical limitation of multimodal SER is that ground-truth transcripts are rarely available in real-world scenarios. Lakomkin et al.~\cite{Lakomkin-IROS-2019} showed that training and evaluating sentiment models with human-transcribed text leads to an optimistic estimate of performance, and argued for end-to-end integration with ASR. Shon et al.~\cite{Lu-Interspeech-2020} subsequently showed that leveraging pre-trained language encoders on ASR transcripts can outperform traditional audio-only sentiment systems, despite recognition errors. Li et al.~\cite{Li-ICASSP-2024} benchmarked SER with ASR transcripts at varying word error rates across multiple datasets, finding that SER is surprisingly tolerant to moderate ASR errors when appropriate fusion and correction mechanisms are employed. Wu et al.~\cite{Liu-AAAI-2024} went one step further and used text-based emotional descriptions generated from non-textual modalities as an additional input to a cross-modal transformer. While these works confirm that automatically generated text is a useful signal for affect modeling, they still require the text branch at inference time. In contrast, we use generated transcripts and translations strictly as privileged information during training, and transfer the resulting knowledge into an audio-only student.
 
\noindent
\textbf{Knowledge distillation for speech and multimodal models.}
Knowledge distillation (KD)~\cite{Hinton-NIPS-2015} is a model compression technique in which a small student network is trained to match the softened output distribution of a larger teacher, thereby inheriting part of the teacher's ``dark knowledge''. A complementary perspective is given by the Learning Using Privileged Information (LUPI) paradigm of Vapnik et al.~\cite{Vapnik-JMLR-2015}, where the teacher has access to additional data (privileged information) that is unavailable at test time. Lopez-Paz et al.~\cite{LopezPaz-ICLR-2016} unified these two perspectives under the framework of generalized distillation, which directly motivates our choice of using automatically generated text as a training-only privileged modality.
 
Building on these foundations, several recent works have applied KD specifically to multimodal emotion recognition. Georgescu et al.~\cite{Georgescu-arXiv-2020} applied a teacher-student distillation objective combining hard labels and softened teacher predictions in an affect recognition setting, which we adopt in our distillation stage. Li et al.~\cite{Li-CVPR-2023} proposed Decoupled Multimodal Distilling, which factorizes multimodal representations into modality-invariant and modality-exclusive spaces before performing cross-modal distillation, improving robustness to modality gaps. Muaz et al.~\cite{Muaz-ARXIV-2024} explicitly targeted the setting where only speech is available at inference time, using a multi-modal teacher (audio, text and video) to supervise an audio-only student via masked training and KD on IEMOCAP \cite{Busso-LRE-2008}. More recently, Li et al.~\cite{Li-AAAI-2025} introduced SUMMER, a heterogeneous multimodal integration framework that uses interactive KD between a unimodal-driven teacher and a multimodal student to mitigate gradient conflicts in multimodal learning. Our work differs from the above methods in two important ways. First, we specifically target sentiment polarity classification on large-scale naturalistic speech (MSP-Podcast \cite{busso2025msp}), rather than acted or conversational corpora such as IEMOCAP \cite{Busso-LRE-2008} and MELD \cite{Poria-ACL-2019}. Second, the privileged modalities in our teacher are not captured by additional sensors, but are automatically generated from the audio itself via ASR and machine translation, which makes the overall pipeline fully audio-driven at inference time, while preserving the benefits of multimodal supervision during training.
 
\vspace{-0.2cm}
\section{Method}
\vspace{-0.1cm}

We propose an audio-centered affect classification framework based on multimodal knowledge distillation. The main objective of this work is to train an efficient speech-only classifier that can benefit from multimodal information during training, while remaining efficient at inference time. To this end, we adopt a teacher-student strategy. The teacher is a multimodal model that jointly processes acoustic and text-based information, whereas the student is an audio-only WavLM \cite{Chen-JSTSP-2022} classifier. The student is optimized using both ground-truth labels and the soft predictions of the teacher, so that part of the multimodal knowledge can be transferred to a unimodal model.

\noindent
\textbf{Multimodal teacher model.}
The teacher model is built to integrate complementary information extracted from speech and text. For each audio sample, we first generate an English transcript using a Faster-Whisper \cite{faster-whisper} automatic speech recognition model. The English transcript is then translated with NLLB-200 \cite{nllb-200-distilled-600M} into additional languages supported by the framework, such as Spanish, German, and French. In this way, the system constructs several textual modalities of the same utterance, each potentially capturing different semantic nuances.

Each active modality is encoded with a dedicated pre-trained backbone. The acoustic branch relies on WavLM-base-plus \cite{Chen-JSTSP-2022}, while the textual branches are modeled with language-specific transformer encoders: RoBERTa-base \cite{twitter-roberta-base-sentiment-latest} for English, RoBERTuito-base \cite{robertuito-sentiment-analysis} for Spanish, GBERT-base \cite{gbert-base} for German, and CamemBERT-base \cite{camembert-base} for French. These encoders produce fixed-dimensional embeddings, which are then projected into a common latent space by modality-specific adapters. Each adapter transforms the corresponding embedding into a fixed number of learnable patch tokens, ensuring that all modalities share the same token dimensionality and the same number of tokens.

The tokens produced by all active modalities are concatenated and passed to a Cascaded Cross-Modal Transformer (CCMT) \cite{Ristea-AIRE-2024}. To preserve modality identity, the model uses modality-specific positional embeddings. CCMT performs a sequence of cross-modal attention operations, progressively integrating acoustic and linguistic information. Each cross-modal attention block combines two modalities. Formally, for any two modalities $i$ and $j$, the output of the cross-attention layer, denoted as $O_{i+j} \in \mathbb{R}^{k \times d}$, is computed as:
\begin{equation}
\label{eq1}
O_{i+j} = \mathrm{softmax}\left( \frac{Q_{i}\cdot K_{j}^{\top}}{\sqrt{d}} \right) \cdot V_{j},
\end{equation}
where $Q_i = T_i \cdot W_Q$, $K = T_j \cdot W_K$, $V = T_j \cdot W_V$, and $d$ represents the token dimension of the attention head. $W_Q, W_K, W_V$ represent learnable weights, while $T_i$ and $T_j$ represent the tokens coming from modalities $i$ and $j$, respectively. The cross-attention layer integrates query tokens from modality $i$ with key and value tokens from modality $j$. The \texttt{[CLS]} token from the final multimodal representation is fed into a classification head that outputs one of the three target classes: \textit{negative}, \textit{neutral}, or \textit{positive}.

The purpose of the teacher is twofold. First, it aims to achieve strong multimodal classification performance by jointly exploiting speech and language information. Second, it provides informative soft targets for the student during distillation, enabling the final audio-only model to benefit from knowledge that is otherwise unavailable at inference time.

\noindent
\textbf{Teacher training.}
The teacher training pipeline is organized into two stages. In the first stage, each backbone is trained independently on the target sentiment classification task. The text encoders are trained on their corresponding textual inputs, while the WavLM model is trained directly on raw audio. This step allows each encoder to learn modality-specific discriminative features before multimodal fusion is performed. For each branch, the best checkpoint is selected based on validation performance.

In the second stage, the selected backbone checkpoints are reused as feature extractors for the multimodal model. Instead of repeatedly forwarding raw inputs through all encoders during CCMT training, we first generate and cache precomputed embeddings for each dataset partition. This reduces the computational cost of multimodal training and makes it easier to compare multiple combinations of modalities under the same setting.

Once the embeddings are available, each modality is passed through its dedicated adapter, which maps the unimodal representation to a sequence of patch tokens in a shared latent space. The token sequences are subsequently fed into the CCMT. While training the multimodal fusion module based on CCMT, the backbone representations remain fixed, through the use of precomputed embeddings. This strategy stabilizes optimization, since the fusion model receives already informative unimodal features. It also accelerates training by removing repeated backbone computations.

After training several multimodal configurations, the best-performing CCMT model is selected and frozen. This model becomes the teacher network used in the distillation stage.

\noindent
\textbf{Audio-only student model.}
The student represents the final deployment model of the proposed framework. It consists of a WavLM backbone followed by a lightweight classification head, and it receives only the raw audio waveform as input. Unlike the teacher, the student does not require transcription, translation, or multimodal fusion during inference, which makes it considerably more efficient. Although the student operates only on the acoustic modality, it is trained to approximate both the hard labels and the behavior of the multimodal teacher. Therefore, the final student can compress part of the cross-modal knowledge of the teacher model into an audio-only model.

\noindent
\textbf{Distillation stage.}
The distillation stage is the core mechanism through which multimodal knowledge is transferred to the student \cite{Hinton-NIPS-2015}. Let $z_t$ and $z_s$ denote the logits produced by the teacher and student models, respectively. We compute softened probability distributions using a temperature parameter $\tau$:
\begin{equation}
p_t = \mathrm{softmax}\left(\frac{z_t}{\tau}\right), \qquad
p_s = \mathrm{softmax}\left(\frac{z_s}{\tau}\right).  
\end{equation}
The student is trained with a combination of two objectives. The first objective is the standard cross-entropy loss with respect to the ground-truth label $y$ (given as a one-hot encoding):
\begin{equation}\label{eq_loss_ce}
   \mathcal{L}_{\mathrm{CE}}(y, p_s) = - \sum_{i=1}^k y_{i} \cdot \log(p_{s,i}),
\end{equation}
where $k$ is the number of classes. The second objective is a distillation loss that encourages the student to match the teacher's softened output distribution:
\begin{equation}
\mathcal{L}_{\mathrm{KD}} = \tau^2 \cdot D_\mathrm{KL}(p_t \,\|\, p_s),
\end{equation}
where $D_\mathrm{KL}$ is the Kullback–Leibler (KL) divergence.

The final training objective is given by:
\begin{equation}\label{eq_L_total}
\mathcal{L} = (1 - \lambda)\cdot \mathcal{L}_{\mathrm{CE}} + \lambda\cdot \mathcal{L}_{\mathrm{KD}},
\end{equation}
where $\lambda \in [0,1]$ controls the balance between hard supervision and teacher guidance.
Through this objective, the student learns not only the correct class assignments, but also the relative confidence structure encoded by the multimodal teacher. This is particularly important in sentiment classification, where class boundaries may be ambiguous and soft predictions can convey useful information about inter-class similarity. As a result, the student can approximate part of the richer decision process of the teacher, despite only having access to audio information.

% \noindent
% \textbf{Inference.}
% At inference time, only the student model is retained. A speech sample is passed directly through the student WavLM encoder and its classification head to generate the final prediction. Therefore, the proposed framework preserves the benefits of multimodal supervision during training, while keeping the inference pipeline fully audio-based and computationally efficient.

\vspace{-0.2cm}
\section{Experiments}
\vspace{-0.1cm}

\noindent
\textbf{Dataset.}
We conduct experiments on the large-scale MSP-Podcast corpus \cite{busso2025msp}, a benchmark dataset for speech-based affect analysis. The experiments follow the official corpus partitions, namely \emph{train} (169,190 samples), \emph{development} (34,399 samples), and \emph{test-1} (46,294 samples), in order to ensure a consistent evaluation protocol.

For the sentiment polarity classification task, the original emotion annotations are mapped to three target classes: \emph{negative}, \emph{neutral}, and \emph{positive}. Negative emotions such as anger, sadness, disgust, contempt, and fear are grouped into the \emph{negative} class, while happy and neutral samples are assigned to \emph{positive} and \emph{neutral} classes, respectively. The remaining samples are assigned according to their valence score, with low-valence utterances mapped to \emph{negative}, and high-valence utterances mapped to \emph{positive}.

During preprocessing, all audio signals are resampled to 16 \texttt{kHz}, while transcript files are aligned at the utterance level through the corresponding file identifiers. This setup enables both multimodal teacher training and audio-only student evaluation within the same experimental framework.

\noindent
\textbf{Implementation details.}
In addition to the available audio recordings, we construct auxiliary textual modalities for multimodal training. Each utterance is first transcribed into English using an ASR model, and the resulting transcript is then translated into additional languages, namely Spanish, German, and French. Thus, each sample can be represented by its acoustic signal together with one or more text-based views derived from speech.

Our experiments rely on a progressive three-step training setup. First, we independently fine-tune the text-only classifiers for each language (English, Spanish, German, and French), and a standard WavLM-base-plus audio baseline (without KD). Second, we leverage these fine-tuned text and audio models as robust feature extractors to train the multimodal teacher model (CCMT). Finally, we establish our deployment-ready pipeline, where a fresh WavLM-base-plus student model learns from the fully-trained CCMT teacher via KD. We implemented all models in PyTorch, utilizing mixed-precision training for the audio and multimodal networks to optimize memory and computational efficiency.

\noindent
\textbf{Hyperparameter setup.}
In Table~\ref{tab:hyperparams}, we report the main training and architectural hyperparameters for all models. To efficiently fine-tune the pre-trained unimodal models, and prevent them from forgetting their original knowledge, we use Low-Rank Adaptation (LoRA) \cite{Hu-ICLR-2022}, while keeping the original pre-trained weights frozen. The text-only classifiers are trained via AdamW for $5$ epochs with a batch size of $256$, an initial learning rate of $2 \cdot 10^{-4}$, a LoRA rank ($r$) of $16$, and a scaling factor ($\alpha$) of $32$, without gradient accumulation. Similarly, the WavLM audio baseline is fine-tuned via AdamW for $5$ epochs with a learning rate of $3 \cdot 10^{-4}$. The LoRA parameters are set to $r=8$ and $\alpha=16$. WavLM uses a physical batch size of $16$ and two gradient accumulation steps.

For the multimodal classification task, the Cascaded Cross-Modal Transformer (CCMT) fuses the linguistic and acoustic representations extracted by the previously fine-tuned unimodal backbones. The CCMT is trained for up to $50$ epochs with the AdamW optimizer and a linear warm-up scheduler. The initial learning rate is $10^{-4}$, the batch size is $128$, and gradient accumulation is turned off. To prevent overfitting, we use an early stopping mechanism that monitors the validation macro-$F_1$ score and stops the training if the performance does not improve.

During the distillation stage, the WavLM student uses the same architecture and training parameters as the audio baseline. However, the student is trained using two combined objectives: the standard cross-entropy loss on the ground-truth labels and a KD loss based on the soft probabilities predicted by the teacher. %This combined loss acts as a regularization technique, helping the student learn the richer relationships between classes captured by the teacher. 
We consider two teacher models for KD, one that combines audio with English and French text, and one that combines audio with English and German text. We use a distillation temperature of $\tau = 2$, and a default KD weight of $\lambda = 0.7$.
To test how the student reacts to the teacher's influence, we also conduct an ablation study on $\lambda$.

% $\$\lambda \in \{0.5, 0.8, 0.9\}$. 

\begin{table}[t]
\centering
\caption{Summary of training and architectural hyperparameters across model configurations.}
\label{tab:hyperparams}
\begin{tabular*}{\hsize}{@{\extracolsep{\fill}}lccc@{}}
\toprule
{Hyperparameter} & {Text Backbones} & {WavLM Baseline \& Student} & {CCMT Teacher} \\
\midrule
Epochs & 5 & 5 & 50 (early stopping) \\
Learning Rate & $2 \cdot 10^{-4}$ & $3 \cdot 10^{-4}$ & $1 \cdot 10^{-4}$ \\
Optimizer & AdamW & AdamW & AdamW \\
Scheduler & Linear Warmup & Linear Warmup & Linear Warmup \\
Batch Size & 256 & 16 & 128 \\
Gradient Accumulation & 1 & 2 & 1 \\
Mixed Precision & False & True & True \\
LoRA Rank ($r$) & 16 & 8 & - \\
LoRA Alpha ($\alpha$) & 32 & 16 & - \\
\midrule
\multicolumn{4}{l}{\emph{Knowledge distillation (student only):}} \\
\midrule

Temperature ($\tau$) & - & 2.0 & - \\
KD Weight ($\alpha$) & - & 0.7 & - \\
\bottomrule
\end{tabular*}
\vspace{-0.6cm}
\end{table}

\noindent
\textbf{Evaluation measures.}
We assess models in terms of accuracy rate and macro-$F_1$. Accuracy assigns equal importance to each test sample, ignoring the class distribution, while macro-$F_1$ computes the average $F_1$ scores over all three classes, treating all classes as equally important.

\noindent
\textbf{Baselines.}
To compare our final unimodal (audio-only) students, we consider two strong audio-based models as baselines. As the first unimodal audio baseline, we consider the fine-tuned WavLM without KD. Having the same architecture as our students, any performance difference between the baseline WavLM and the WavLM-based students is due to knowledge distillation.
As the second unimodal audio baseline, we fine-tune the Whisper-base encoder~\cite{Radford-ICML-2023} (74M parameters) for three-class sentiment classification on the MSP-Podcast corpus. We introduce this baseline to better assess the capabilities of the WavLM architecture w.r.t.~direct unimodal competitors.

\begin{table}[t]
\caption{Performance of unimodal, multimodal and KD-based models on the MSP-Podcast corpus \cite{busso2025msp}, alongside inference time in \texttt{ms} (measured on one Nvidia RTX 4060 8GB GPU, using a batch size of 8).}
\label{tab:main_res}
\begin{tabular*}{\hsize}{@{\extracolsep{\fill}}lccccc ccc@{}}
\toprule
\multirow{2.5}{*}{Model} & \multirow{2.5}{*}{Audio} & \multicolumn{4}{c}{Text} & \multirow{1.5}{*}{Macro} & \multirow{2.5}{*}{Accuracy} & \multirow{1.5}{*}{Time} \\
\cmidrule{3-6}
 &  & En & Es & De & Fr & $F_1$ &  & (ms) \\
\midrule
Whisper & \cmark & \xmark & \xmark & \xmark & \xmark & 0.5699 & 0.5723 & 180 \\
WavLM & \cmark & \xmark & \xmark & \xmark & \xmark & 0.6239 & 0.6425 & 213 \\
\midrule
RoBERTa & \xmark & \cmark & \xmark & \xmark & \xmark & 0.6102 & 0.6235 & 27 \\
RoBERTuito & \xmark & \xmark & \cmark & \xmark & \xmark  & 0.5751 & 0.5981 & 138 \\
GBERT & \xmark & \xmark & \xmark & \cmark & \xmark & 0.5742 & 0.5893 & 139 \\
CamemBERT & \xmark & \xmark & \xmark & \xmark & \cmark  & 0.5706 & 0.5867 & 138 \\
\midrule
CCMT & \cmark & \cmark & \xmark & \xmark & \xmark & 0.6812 & 0.6910 & 3404 \\
CCMT & \cmark & \cmark & \cmark & \xmark & \xmark & 0.6793 & 0.6881 & 30112 \\
CCMT & \cmark & \cmark & \xmark & \cmark & \xmark & 0.6828 & 0.6924 & 27714 \\
CCMT & \cmark & \cmark & \xmark & \xmark & \cmark & 0.6826 & 0.6940 & 25082 \\
CCMT & \cmark & \cmark & \cmark & \cmark & \cmark & 0.6812 & 0.6903 & 76000 \\
\midrule
WavLM (KD from Audio + En + Fr CCMT) & \cmark & \xmark & \xmark & \xmark & \xmark  & 0.6393 & 0.6506 & 211 \\
WavLM (KD from Audio + En + De CCMT) & \cmark & \xmark & \xmark & \xmark & \xmark  & 0.6329 & 0.6474 & 210 \\
\bottomrule
\end{tabular*}
\vspace{-0.5cm}
\end{table}                                               

\noindent
\textbf{Main results and ablations.} In Table \ref{tab:main_res}, we report the results for unimodal, multimodal and KD-based models on the MSP-Podcast corpus \cite{busso2025msp}. Among the two audio-only baselines, WavLM obtains clearly better performance, surpassing Whisper by $+7\%$ in terms of accuracy. This also suggests that further improving the already strong WavLM is not going to be easy. Taken individually, the text-only models obtain lower performance levels than the audio-based WavLM. This is an expected phenomenon, since text-only models rely on automatically generated inputs that contain inherent errors, thereby degrading performance. Among text-only models, RoBERTa obtains the highest performance levels, likely because text translations to Spanish, German and French contain additional errors introduced during the NMT process, which further degrades performance. Yet, an encouraging observation is that all text-only models outperform the audio-based Whisper, which indicates that these models may still be useful in a multimodal context. Looking at the multimodal (audio-text) CCMT variants, we observe that integrating multiple modalities leads to significant performance boosts over the audio-only WavLM model. Performance gains are around $+5\%$ in absolute percentage points, which is quite impressive, given that all text modalities are generated via ASR and NMT. KD from CCMT to WavLM is applied for the top two CCMT variants. Both WavLM students outperform the fine-tuned WavLM baseline, suggesting that distillation is useful. Actually, the utility of KD becomes more evident when we analyze the computational time. While CCMT achieves significant performance gains, all CCMT variants are slower by at least one order of magnitude than the audio-only models. In contrast, the WavLM students attain the same inference speed as the audio-only models, since they do not require additional text-based inputs during inference.

In summary, our empirical results indicate that generated transcripts and translations provide additional sentiment polarity clues that can be effectively harnessed by a multimodal model. Moreover, the results confirm that knowledge from heavy multimodal models can be distilled into efficient unimodal students.

\begin{figure}[t]
    \centering
    \includegraphics[width=0.52\linewidth, trim={0cm 0cm 0cm 1.8cm}, clip]{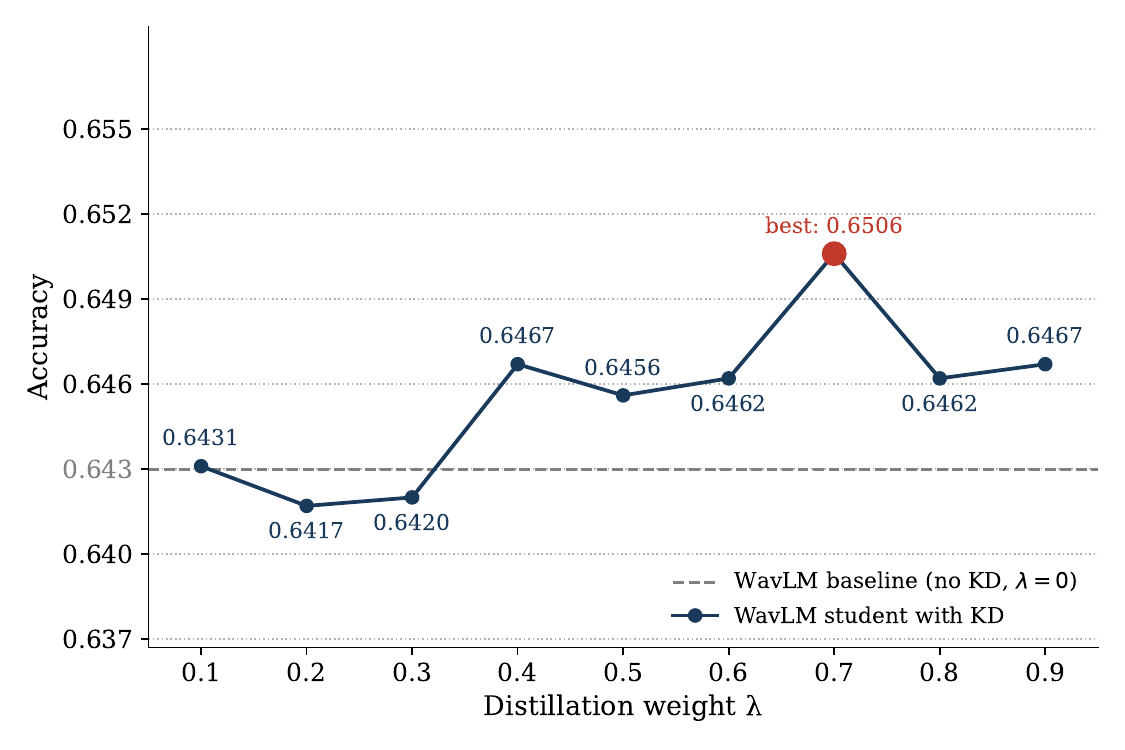}
    \vspace{-0.3cm}
    \caption{Accuracy rates of the WavLM student across different values for the distillation weight $\lambda$. The teacher is the CCMT based on Audio + En + Fr. The dashed line marks the WavLM baseline trained without KD ($\lambda=0$). The best result is obtained at $\lambda=0.7$.}
    \label{fig:lambda_sweep}
    \vspace{-0.3cm}
\end{figure}

% \begin{figure}[htbp]
%     \centering
    
%     % Prima imagine: WavLM Audio
%     \subfloat[WavLM (Baseline)\label{fig:cm_wavlm}]{%
%         \includegraphics[width=0.32\textwidth]{wavlm_audio_confusion_matrix.pdf}%
%     }\hfill
%     % A doua imagine: CCMT Multimodal
%     \subfloat[CCMT (Teacher)\label{fig:cm_ccmt}]{%
%         \includegraphics[width=0.32\textwidth]{ccmt_multimodal_text_en_text_fr_audio_confusion_matrix.pdf}%
%     }\hfill
%     % A treia imagine: WavLM KD
%     \subfloat[WavLM KD ($\alpha=0.7$)\label{fig:cm_wavlm_kd}]{%
%         \includegraphics[width=0.32\textwidth]{wavlm_audio_kd_text_en_text_fr_audio_a07_confusion_matrix.pdf}%
%     }
    
%     \caption{Confusion matrices for the baseline WavLM, the multimodal CCMT teacher, and the distilled WavLM KD model.}
%     \label{fig:confusion_matrices}
% \end{figure}

\noindent
\textbf{Ablation of $\lambda$.} In Figure \ref{fig:lambda_sweep}, we present KD results with various values for $\lambda$, the hyperparameter that controls the influence of the two loss components in Eq.~\eqref{eq_L_total}. When $\lambda=0$, the KD process is turned off, so the results correspond to the WavLM model without KD. The student WavLM model surpasses the WavLM (no KD) baseline, when the value chosen for $\lambda \geq 0.4$. This indicates that multimodal information is generally beneficial during KD, providing performance gains, even when $\lambda$ is not optimally configured.

\vspace{-0.2cm}
\section{Conclusion}
\vspace{-0.1cm}

In this work, we proposed a novel knowledge distillation framework for audio sentiment polarity classification, where a multimodal teacher integrates the speech signal with automatically generated text transcripts and translations through a cascaded cross-modal transformer architecture, and transfers its knowledge to an efficient audio-only student. Experiments on the large-scale MSP-Podcast corpus confirmed both of our working hypotheses. First, augmenting the audio modality with generated multilingual text yielded substantial gains over the unimodal WavLM baseline, with improvements of up to $+5.89\%$ in terms of macro-$F_1$ and $+5.15\%$ in terms of accuracy, while our ablation study indicated that both ASR transcripts and machine-translated texts contribute to the final performance. Second, the multimodal knowledge captured by the teacher was effectively distilled into the audio-only student, boosting its performance by $+1.54\%$ in terms of macro-$F_1$ and $+0.81\%$ in terms of accuracy, without introducing any computational overhead at inference time. Taken together, these results demonstrate that automatically generated textual modalities can serve as effective privileged information for training efficient audio-only sentiment classifiers.

\vspace{-0.3cm}
\section*{Acknowledgments}
\vspace{-0.1cm}

This research is supported by PPC Romania via the 2025-2026 Scholarship Program.

\vspace{-0.1cm}
\bibliography{ref}

@inproceedings{Georgescu-arXiv-2020,
    title = "{Teacher-Student Training and Triplet Loss for Facial Expression Recognition under Occlusion}", 
    author = {Georgescu, Mariana-Iuliana and Ionescu, Radu Tudor},
    booktitle={Proceedings of ICPR}, 
  year={2020},
  pages={2288--2295},
}

@article{Ristea-AIRE-2024,
author = {Ristea, Nicolae-Catalin and Anghel, Andrei and Ionescu, Radu Tudor},
title = "{Cascaded cross-modal transformer for audio-textual classification}",
year = {2024},
journal = {Artificial Intelligence Review},
volume={57},
pages={225},
}

@misc{faster-whisper,
author = {Klein, Guillaume and others},
title = "{faster-whisper: Faster Whisper transcription with CTranslate2}",
year = {2023},
publisher = {GitHub},
howpublished = {\url{https://github.com/SYSTRAN/faster-whisper}}
}

@article{nllb-200-distilled-600M,
  title="{No Language Left Behind: Scaling human-centered machine translation}",
  author={{NLLB Team} and Costa-Juss{\`a}, Marta R. and Cross, James and {\c{C}}elebi, Onur and Elbayad, Maha and Heafield, Kenneth and Heffernan, Kevin and Kalbassi, Elahe and Lam, Janice and Licht, Daniel and Maillard, Jean and others},
  journal={arXiv preprint arXiv:2207.04672},
  year={2022}
}

@inproceedings{twitter-roberta-base-sentiment-latest,
    title = "{T}weet{E}val: Unified Benchmark and Comparative Evaluation for Tweet Classification",
    author = "Barbieri, Francesco  and
      Camacho-Collados, Jose  and
      Espinosa Anke, Luis  and
      Neves, Leonardo",
    booktitle = "Findings of EMNLP",
    year = "2020",
    pages = "1644--1650",
}

@inproceedings{robertuito-sentiment-analysis,
    title = "{R}o{BERT}uito: a pre-trained language model for social media text in {S}panish",
    author = "P{\'e}rez, Juan Manuel  and
      Furman, Dami{\'a}n Ariel  and
      Alonso Alemany, Laura  and
      Luque, Franco M.",
    booktitle = "Proceedings of LREC",
    year = "2022",
     pages = "7235--7243",
}

@inproceedings{gbert-base,
    title = "{G}erman{'}s Next Language Model",
    author = {Chan, Branden  and
      Schweter, Stefan  and
      M{\"o}ller, Timo},
    booktitle = "Proceedings of COLING",
    year = "2020",
    pages = "6788--6796",
}

@inproceedings{camembert-base,
    title = "{C}amem{BERT}: a Tasty {F}rench Language Model",
    author = "Martin, Louis  and
      Muller, Benjamin  and
      Ortiz Su{\'a}rez, Pedro Javier  and
      Dupont, Yoann  and
      Romary, Laurent  and
      de la Clergerie, {\'E}ric  and
      Seddah, Djam{\'e}  and
      Sagot, Beno{\^i}t",
    booktitle = "Proceedings of ACL",
    year = "2020",
    pages = "7203--7219",
}

@article{busso2025msp,
  title="{The MSP-Podcast Corpus}",
  author={Busso, Carlos and Lotfian, Reza and Sridhar, Kusha and Salman, Ali N and Lin, Wei-Cheng and Goncalves, Lucas and Parthasarathy, Srinivas and Naini, Abinay Reddy and Leem, Seong-Gyun and Martinez-Lucas, Luz and others},
  journal={arXiv preprint arXiv:2509.09791},
  year={2025}
}

@article{GEORGE2024127015,
title = {A review on speech emotion recognition: A survey, recent advances, challenges, and the influence of noise},
journal = {Neurocomputing},
volume = {568},
pages = {127015},
year = {2024},
author = {Swapna Mol George and P. Muhamed Ilyas},
}

@article{Lian-Entropy-2023,
AUTHOR = {Lian, Hailun and Lu, Cheng and Li, Sunan and Zhao, Yan and Tang, Chuangao and Zong, Yuan},
TITLE = {A Survey of Deep Learning-Based Multimodal Emotion Recognition: Speech, Text, and Face},
JOURNAL = {Entropy},
VOLUME = {25},
YEAR = {2023},
NUMBER = {10},
pages = {1440},
}

@article{Hsu-TASLP-2021,
author = {Hsu, Wei-Ning and Bolte, Benjamin and Tsai, Yao-Hung Hubert and Lakhotia, Kushal and Salakhutdinov, Ruslan and Mohamed, Abdelrahman},
title = "{HuBERT: Self-Supervised Speech Representation Learning by Masked Prediction of Hidden Units}",
year = {2021},
volume = {29},
journal = {IEEE/ACM Transactions on Audio, Speech and Language Processing},
pages = {3451--3460},
}

@article{Chen-JSTSP-2022,
   title="{WavLM: Large-Scale Self-Supervised Pre-Training for Full Stack Speech Processing}",
   author={Chen, Sanyuan and Wang, Chengyi and Chen, Zhengyang and Wu, Yu and Liu, Shujie and Chen, Zhuo and Li, Jinyu and Kanda, Naoyuki and Yoshioka, Takuya and Xiao, Xiong and Wu, Jian and Zhou, Long and Ren, Shuo and Qian, Yanmin and Qian, Yao and Wu, Jian and Zeng, Michael and Yu, Xiangzhan and Wei, Furu},
  journal={IEEE Journal of Selected Topics in Signal Processing}, 
  year={2022},
  volume={16},
  number={6},
  pages={1505--1518},
}

@article{Wang-ARXIV-2021,
 title="{A Fine-tuned Wav2vec 2.0/HuBERT Benchmark For Speech Emotion Recognition, Speaker Verification and Spoken Language Understanding}", 
  author={Wang, Yingzhi and Boumadane, Abdelmoumene and Heba, Abdelwahab},
  journal={arXiv preprint arXiv:2111.02735},
  year={2021}
}

@inproceedings{Naini-Interspeech-2025,
author = {Naini, Abinay and Goncalves, Lucas and Salman, Ali and Mote, Pravin and \"{U}lgen, Ismail and Thebaud, Thomas and Moro-Vel\'{a}zquez, Laureano and Garcia, Leibny and Dehak, Najim and Sisman, Berrak and Busso, Carlos},
year = {2025},
pages = {4668--4672},
title = "{The Interspeech 2025 Challenge on Speech Emotion Recognition in Naturalistic Conditions}",
booktitle={Proceedings of INTERSPEECH},
}

@inproceedings{Yoon-SLT-2018,
  author={Yoon, Seunghyun and Byun, Seokhyun and Jung, Kyomin},
  booktitle={Proceedings of SLT}, 
  title={Multimodal Speech Emotion Recognition Using Audio and Text}, 
  year={2018},
  pages={112--118},
}

@inproceedings{Tsai-ACL-2019,
  title={Multimodal transformer for unaligned multimodal language sequences},
  author={Tsai, Yao-Hung Hubert and Bai, Shaojie and Liang, Paul Pu and Kolter, J. Zico and Morency, Louis-Philippe and Salakhutdinov, Ruslan},
  booktitle={Proceedings of ACL},
  pages={6558--6569},
  year={2019}
}

@article{Luo-ARXIV-2022,
      title="{ScaleVLAD: Improving Multimodal Sentiment Analysis via Multi-Scale Fusion of Locally Descriptors}", 
      author={Huaishao Luo and Lei Ji and Yanyong Huang and Bin Wang and Shenggong Ji and Tianrui Li},
      year={2021},
     journal={arXiv preprint arXiv:2112.01368},
}

@article{Zhang-ARXIV-2022,
      title={A Self-Adjusting Fusion Representation Learning Model for Unaligned Text-Audio Sequences}, 
      author={Kaicheng Yang and Ruxuan Zhang and Hua Xu and Kai Gao},
  journal={arXiv preprint arXiv:2212.11772},
  year={2022}
}

@article{Vamsidhar-SciRep-2025,
  title   = {Hierarchical cross-modal attention and dual audio pathways for enhanced multimodal sentiment analysis},
  author  = {Vamsidhar, D. and Desai, Parth and Shahade, Aniket K. and Patil, Shruti and Deshmukh, Priyanka V.},
  journal = {Scientific Reports},
  volume  = {15},
  pages   = {25440},
  year    = {2025},
}

@inproceedings{Lakomkin-IROS-2019,
  author={Lakomkin, Egor and Zamani, Mohammad Ali and Weber, Cornelius and Magg, Sven and Wermter, Stefan},
  booktitle={Proceedings of ICRA}, 
  title={Incorporating End-to-End Speech Recognition Models for Sentiment Analysis}, 
  year={2019},
  pages={7976--7982},
}

@inproceedings{Lu-Interspeech-2020,
      title={Leveraging Pre-trained Language Model for Speech Sentiment Analysis}, 
      author={Suwon Shon and Pablo Brusco and Jing Pan and Kyu J. Han and Shinji Watanabe},
       booktitle={Proceedings of INTERSPEECH},
  pages={3420--3424},
  year={2021}
}

@INPROCEEDINGS{Li-ICASSP-2024,
  author={Li, Yuanchao and Bell, Peter and Lai, Catherine},
  booktitle={Proceedings of SLT}, 
  title="{Speech Emotion Recognition With ASR Transcripts: a Comprehensive Study on Word Error Rate and Fusion Techniques}", 
  year={2024},
  pages={518--525},
}

@inproceedings{Liu-AAAI-2024,
  title={Enriching multimodal sentiment analysis through textual emotional descriptions of visual-audio content},
  author={Wu, Sheng and He, Dongxiao and Wang, Xiaobao and Wang, Longbiao and Dang, Jianwu},
  booktitle={Proceedings of AAAI},
  volume={39},
  number={2},
  pages={1601--1609},
  year={2025}
}

@article{Hinton-NIPS-2015,
  title={Distilling the knowledge in a neural network},
  author={Hinton, Geoffrey and Vinyals, Oriol and Dean, Jeff},
  journal={arXiv preprint arXiv:1503.02531},
  year={2015}
}

@article{Vapnik-JMLR-2015,
  author  = {Vladimir Vapnik and Rauf Izmailov},
  title   = {Learning Using Privileged Information: Similarity Control and Knowledge Transfer},
  journal = {Journal of Machine Learning Research},
  year    = {2015},
  volume  = {16},
  number  = {61},
  pages   = {2023--2049},
}

@inproceedings{LopezPaz-ICLR-2016,
  title={Unifying distillation and privileged information},
  author={Lopez-Paz, David and Bottou, L{\'e}on and Sch{\"o}lkopf, Bernhard and Vapnik, Vladimir},
  booktitle={Proceedings of ICLR},
  year={2016}
}

@inproceedings{Li-CVPR-2023,
  title={Decoupled multimodal distilling for emotion recognition},
  author={Li, Yong and Wang, Yuanzhi and Cui, Zhen},
  booktitle={Proceedings of CVPR},
  pages={6631--6640},
  year={2023}
}

@article{Muaz-ARXIV-2024,
  title={Bridging Modalities: Knowledge Distillation and Masked Training for Translating Multi-Modal Emotion Recognition to Uni-Modal, Speech-Only Emotion Recognition},
  author={Muaz, Muhammad and Paull, Nathan and Malagavalli, Jahnavi},
  journal={arXiv preprint arXiv:2401.03000},
  year={2024}
}

@inproceedings{Li-AAAI-2025,
  title={Unimodal-driven distillation in multimodal emotion recognition with dynamic fusion},
  author={Li, Jiagen and Yu, Rui and Huang, Huihao and Yan, Huaicheng},
  booktitle={Proceedings of ICME},
  pages={1--6},
  year={2025},
}

@inproceedings{Luo-AffCon-2019,
  title={Audio Sentiment Analysis by Heterogeneous Signal Features Learned from Utterance-Based Parallel Neural Network},
  author={Luo, Ziqian and Xu, Hua and Chen, Feiyang},
  booktitle={Proceedings of AffCon@AAAI},
  pages={80--87},
  year={2019},
}

@article{Atmaja-Sen-2022,
  title={Sentiment analysis and emotion recognition from speech using universal speech representations},
  author={Atmaja, Bagus Tris and Sasou, Akira},
  journal={Sensors},
  volume={22},
  number={17},
  pages={6369},
  year={2022},
}

@incollection{Mohanty-ADMTMSC-2022,
  title={Verbal sentiment analysis and detection using recurrent neural network},
  author={Mohanty, Mohan Debarchan and Mohanty, Mihir Narayan},
  booktitle={Advanced Data Mining Tools and Methods for Social Computing},
  pages={85--106},
  year={2022},
  publisher = {Academic Press},
}

@inproceedings{Shruti-BigComp-2023,
  title="{A comparative study on Bengali speech sentiment analysis based on audio data}",
  author={Shruti, Abanti Chakraborty and Rifat, Rakib Hossain and Kamal, Marufa and Alam, Md Golam Rabiul},
  booktitle={Proceedings of BigComp},
  pages={219--226},
  year={2023},
}

@inproceedings{Garcia-ICTC-2024,
  title="{Optimized Sentiment Analysis in Tagalog Speech Using PCA and BRNN on Prosodic Suprasegmental and MFCC Features}",
  author={Garcia, Ailen B. and Gerardo, Bobby D. and Medina, Ruji P.},
  booktitle={Proceedings of ICTC},
  pages={60--65},
  year={2024},
}

@inproceedings{Bulkrock-ICTCS-2025,
  title="{Sentiment Analysis of Customer Feedback and Reviews in E-Commerce Systems}",
  author={Bulkrock, Olla and Qusef, Abdallah and BaniMustafa, Ahmed},
  booktitle={Proceedings of ICTCS},
  pages={379--385},
  year={2025},
}

@article{Anilsagar-JSEE-2025,
  title="{The evolution of sentiment analysis and conversational AI: Techniques applications and future research directions}",
  author={Anilsagar, T. and Syed, S. Syed Abdul},
  journal={Journal of Systems Engineering and Electronics},
  volume={35},
  number={1},
  pages={71--82},
  year={2025}
}

@article{Zhang-TAC-2026,
  title={Personality-Aware Multimodal Driver Emotion Recognition Towards Intelligent Connected Vehicles},
  author={Zhang, Puning and Hu, Mengxue and Zhang, Hongbin and Wu, Chao and Yang, Zhigang},
  journal={IEEE Transactions on Affective Computing},
year={2026},
  volume={17},
  number={1},
  pages={801--816},

}

@article{Shanthi-HTL-2025,
  title={An integrated approach for mental health assessment using emotion analysis and scales},
  author={Shanthi, N. and Stonier, Albert Alexander and Sherine, Anli and Devaraju, T. and Abinash, S. and Ajay, R. and Arul Prasath, V. and Ganji, Vivekananda},
  journal={Healthcare Technology Letters},
  volume={12},
  number={1},
  pages={e12040},
  year={2025},
}

@article{Liu-EC-2025,
  title="{Development of interactive English e-learning video entertainment teaching environment based on virtual reality and game teaching emotion analysis}",
  author={Liu, Jia},
  journal={Entertainment Computing},
  volume={52},
  pages={100884},
  year={2025},
  }

@inproceedings{Radford-ICML-2023,
  title="{Robust Speech Recognition via Large-Scale Weak Supervision}",
  author={Radford, Alec and Kim, Jong Wook and Xu, Tao and Brockman, Greg and McLeavey, Christine and Sutskever, Ilya},
  booktitle={Proceedings of ICML},
  pages={28492--28518},
  year={2023},
}

@inproceedings{Baevski-NeurIPS-2020,
  title={{wav2vec} 2.0: A framework for self-supervised learning of speech representations},
  author={Baevski, Alexei and Zhou, Yuhao and Mohamed, Abdelrahman and Auli, Michael},
  booktitle={Proceedings of NeurIPS},
  volume={33},
  pages={12449--12460},
  year={2020},
}

@article{Pratap-JMLR-2024,
  title={Scaling Speech Technology to 1,000+ Languages},
  author={Vineel Pratap and Andros Tjandra and Bowen Shi and Paden Tomasello and Arun Babu and Sayani Kundu and Ali Elkahky and Zhaoheng Ni and Apoorv Vyas and Maryam Fazel-Zarandi and Alexei Baevski and Yossi Adi and Xiaohui Zhang and Wei-Ning Hsu and Alexis Conneau and Michael Auli},
  journal = {Journal of Machine Learning Research},
  year    = {2024},
  volume  = {25},
  number  = {97},
  pages   = {1--52},
}

@inproceedings{Kim-ICML-2021,
      title="{Conditional Variational Autoencoder with Adversarial Learning for End-to-End Text-to-Speech}", 
      author={Jaehyeon Kim and Jungil Kong and Juhee Son},
      year={2021},
  booktitle={Proceedings of ICML},
pages={5530--5540},
}

@inproceedings{Ristea-INTERSPEECH-2022,
  title="{SepTr: Separable Transformer for Audio Spectrogram Processing}",
  author={Ristea, Nicolae-Cătălin and Ionescu, Radu Tudor and Khan, Fahad Shahbaz},
  year={2022},
  booktitle={Proceedings of INTERSPEECH},
  pages={4103--4107},
}

@inproceedings{Gong-INTERSPEECH-2021,
  title="{{AST}: Audio Spectrogram Transformer}",
  author={Gong, Yuan and Chung, Yu-An and Glass, James},
  booktitle={Proceedings of INTERSPEECH},
  pages={571--575},
  year={2021},
}

@inproceedings{ Devlin-NAACL-2019,
  title="{BERT: Pre-training of deep bidirectional transformers for language understanding}",
  author={Devlin, Jacob and Chang, Ming-Wei and Lee, Kenton and Toutanova, Kristina},
  booktitle={Proceedings of NAACT-HLT},
  pages={4171--4186},
  year={2019}
}

@inproceedings{Menadil-ICPR-2024,
  title={Learning Using Generated Privileged Information by Text-to-Image Diffusion Models},
  author={Menadil, Rafael-Edy and Georgescu, Mariana-Iuliana and Ionescu, Radu Tudor},
  booktitle={Proceedings of ICPR},
  pages={423--438},
  year={2024},
}

@inproceedings{Ristea-ACMMM-2023,
  title={Cascaded cross-modal transformer for request and complaint detection},
  author={Ristea, Nicolae-Catalin and Ionescu, Radu Tudor},
  booktitle={Proceedings of ACMMM},
  pages={9467--9471},
  year={2023}
}

@article{Busso-LRE-2008,
  title="{IEMOCAP: Interactive emotional dyadic motion capture database}",
  author={Busso, Carlos and Bulut, Murtaza and Lee, Chi-Chun and Kazemzadeh, Abe and Mower, Emily and Kim, Samuel and Chang, Jeannette N. and Lee, Sungbok and Narayanan, Shrikanth S.},
  journal={Language Resources and Evaluation},
  volume={42},
  number={4},
  pages={335--359},
  year={2008},
}

@inproceedings{Poria-ACL-2019,
    title = "{MELD: A Multimodal Multi-Party Dataset for Emotion Recognition in Conversations}",
    author = "Poria, Soujanya  and
      Hazarika, Devamanyu  and
      Majumder, Navonil  and
      Naik, Gautam  and
      Cambria, Erik  and
      Mihalcea, Rada",
    booktitle = "Proceedings of ACL",
    year = "2019",
    pages = "527--536",
}

@inproceedings{Hu-ICLR-2022,
author = {Hu, Edward J. and Shen, Yelong and Wallis, Phillip and Allen-Zhu, Zeyuan and Li, Yuanzhi and Wang, Shean and Wang, Lu and Chen, Weizhu},
title = "{LoRA: Low-Rank Adaptation of Large Language Models}",
year = {2022},
booktitle = {Proceedings of ICLR},
}
\bibliographystyle{elsarticle-harv}

\end{document}